# Robotics Evolution: from Remote Brain to Cloud

Alaa F. Sheta[1], Nazeeh Ghatasheh[2*], Hossam Faris[3] and Ali Rodan[3]

[1]Department of Computing Sciences, Texas A&M University-Corpus Christi, TX, USA

[2]Faculty of Information Technology and Systems, The University of Jordan, Aqaba, Jordan.

[3]King Abdullah II School for Information Technology, The University of Jordan, Amman, Jordan.

alaa.sheta@tamucc.edu, [n.ghatasheh, hossam.faris, a.rodan]@ju.edu.jo

***Abstract***

*Robotic systems have been evolving since decades and touching almost all aspects of life, either for leisure or critical applications. Most of traditional robotic systems operate in well-defined environments utilizing pre-configured on-board processing units. However, modern and foreseen robotic applications ask for complex processing requirements that exceed the limits of on-board computing power. Cloud computing and the related technologies have high potential to overcome on-board hardware restrictions and can improve the performance efficiency. This research highlights the advancements in robotic systems with focus on cloud robotics as an emerging trend. There exists an extensive amount of effort to leverage the potentials of robotic systems and to handle arising shortcomings. Moreover, there are promising insights for future breed of intelligent, flexible, and autonomous robotic systems in the Internet of Things era.*

## 1. Introduction

For more than five decades, robots have been successfully used to replace human in doing dangerous and tedious work, including hazard environments. They have been used in manufacturing [1, 2], health care [3–5], defense [6, 7], space operations [8], classroom [9], education [10] and many other applications. The utilization of robots improved the quality of life, reduced risks, and expedited processing time. For example, enhancing the productivity in manufacturing industry, lever- aging the accuracy of critical surgeries, and advanced battle planning for military operations [11–13].

A typical robot consists of sensors, control system, and actuators. Sensors are in charge of gathering information about an environment. The robot uses the the collected information to manage its behavior and take appropriate action based on each situation. There are various types of sensors, for example light sensor, laser sensor, ultrasound transceiver, microphone, and camera. The robot control system, alternately robot brain [14], determines the behavior of the robot under various operating conditions. The brain is a program that receives continuous flow of inputs from the sensors, processes the inputs, and consequently sends appropriate action commands to the actuators. The actuators are mounted on the robot body and responsible for translating action commands into physical activities in the environment.

Embedding the robot's control program on its micro-controller is a common practice. Therefore, all necessary computations and data reside on the mother- board of the robot itself. This configuration limits the capabilities of the robot with respect to computation power, memory, storage, and power consumption.

The research articles of M. Inaba [14–18] introduced the term "Remote-Brained Robots". The idea of Inaba suggests implementing a remote brain of the robot apart from its body. The possible characteristics and configurations of a remote brain are unlimited, such possibilities enable reconsidering more powerful robot brains. Other research area in remote brain can be found in [19, 20]. Accordingly, wireless communication technologies play an essential role in data communication for the remote-brained robots.

This paper presents earlier applications of remote-brained robots, discussion on the advancements in robotic technology towards networked robot systems, and fi- nally the rise of cloud robots. The presentation of the ideas is organized as follows: Section 2, presents research work on remote-brained robots and their utilization to solve challenging problems such as landmine detection. Networked robots chal- lenges and applications will be presented in Section 3. The rise of cloud robots, their functions and main advantages will be discussed in Section 5. Future research directions and applications are presented in Section 6.

## 2. Remote-Brained Robots

A registered patent of robot aircraft back in 1958 illustrates the early works to- wards more advanced remote controlled/brained robots [21]. Compared to current advancements and efforts in the field the proposed platform in [21] is primitive.

Early research work on remote-brained robots in [15] illustrates the use of wire- less communication network for the communication between a heavy brain and the body of a robot. The authors studied number of configurations to develop an appropriate control method. In which a video screen is able to present touch, force, and other visual data over the sensor image. They claimed that their approach enabled building a massively powerful parallel computing base alongside the real time vision system. A robot with the brain separated from the body is presented in [16]. The brain resides in what they called "the mother environment" and the robot communicates with it using radio links.

An interesting research theme in [22] aims to conceptualize a platform for future brained-robots. The authors emphasized the idea of separating physically and logically the robot body and its "brain". Such separation makes it more encouraging and possible to utilize advanced and powerful processing techniques. Further- more, the authors conducted several experiments to perform different tasks using different types of robots that share the same remote intelligence platform. The experimented robots were motion enabled, vision-based and others. They concluded that a real world application of the computer intelligence needs more investigation and research effort. Therefore their platform allows students/researchers to experiment several algorithms and robots aiming to enrich the field with more applicable solutions and gap filling.

In [23] the authors developed and implemented a large scale software platform for real time robotic systems. The platform consists of three layers that are the Mother, Brain, and Sensor-Motor. The authors proposed a mother of tools in charge of producing programs to control the brain. In addition, a method to develop variety of brain architectures with flexible multiprocess network.

Remote-brained robots industry faces many challenges. One of the main challenges is the real-time integration of a large-scale brain and a lightweight humanoid body. The research project in [17] tried to prove that remote-brained robots would overcome main challenges of the industry.

A real remote-brained robot that has human-machine interaction, vision systems, and manipulator was illustrated in [24]. The brain was separated from the robot's body. The brain was left on the mother environment, which has the brain's software. The brain and body are interacting via wireless communication links. Among several examples of remote-brained robots there are "Vision-Equipped Apelike Robot" [16], "Remote-brained LEGO robot" [25]. Table 1 lists part of the research work in the area of remote-brained robotics

**Table 1. Part of Remote Brained Robotics Research**

| Area/Subject | Objective | Ref. |
|---|---|---|
| Real time vision and parallel computing | Developing a control method and building a parallel computing base. | [15] |
| Vision-Equipped Apelike Robot | Developing a remote-brained robot and utilizing radio links for communication. | [16] |
| Real Time Robotics | Developing and implementing a large scale software platform and tools. | [23] |
| Project Work | Illustrating how robots could overcome challenges. | [17] |
| Human-Machine Interaction | Developing a real interactive remote-brained robot. | [24] |
| Robotics Platform | Developing futuristic platform with a shared intelligence base for different types of robots. | [22] |
| Registered Patent | Defining a robot aircraft (1958) | [21] |
| Minesweepers | Building Remote-brained LEGO robot. | [25] |
| Robotics Laboratory | Developing various types of robots. | [14,16,26, 27] |
| … | | |

**2.1 Vision-Equipped Apelike Robot**

The Johou Systems Kougaku (JSK) Laboratory has been developing various types of robots since the early 90's [14, 16, 26, 27]. The main categories of JSK robots are wheeled, quadruped, and humanoid. Vision-Equipped Apelike Robot [16] is one of their various "remote-brained" robots. The main benefit from Apelike robot is to study the "vision-based behaviors" of locomotion. It enables the control of an equipped camera at any point of view, the mobility in unstructured environments, and the handling of objects using robot arms.

Apelike robot has a remote brain that makes it able to perform complex vision-based operations. A major physical characteristic of the robot is its light weight, which is possible because the processing system is not part of the physical robot. Figure 1 illustrates the main components of the Apelike robot system.

The system is composed of three main components:

(i) The Remote Brain which is powered by Unix workstation on which the Euslisp language is used to model the behaviors of robot. The object-oriented modeling language made it possible to design various behaviors and interactions. Therefore, the remote brain is used to guide and control the balance, motion, and interactions with the physical robot.

(ii) The Brain-Body Interface which is equipped with a transputer for interpretation and communication. It consists of two sub-transputers: a) Vision re- ceiver and interpreter, and b) Motion interpreter and transmitter. The interface is used to process a relatively adequate amount of images for motion detection.

(iii) The Physical Robot Body weighs about 2.5 kilograms including power supply (i.e. batteries). The body is equipped with two main radio-based communication components; the first one is the receiver of motion signals and the second is the transmitter of sensors data. The encoded communication between the robot and the brain enables it to perform kinetic motion behaviors.

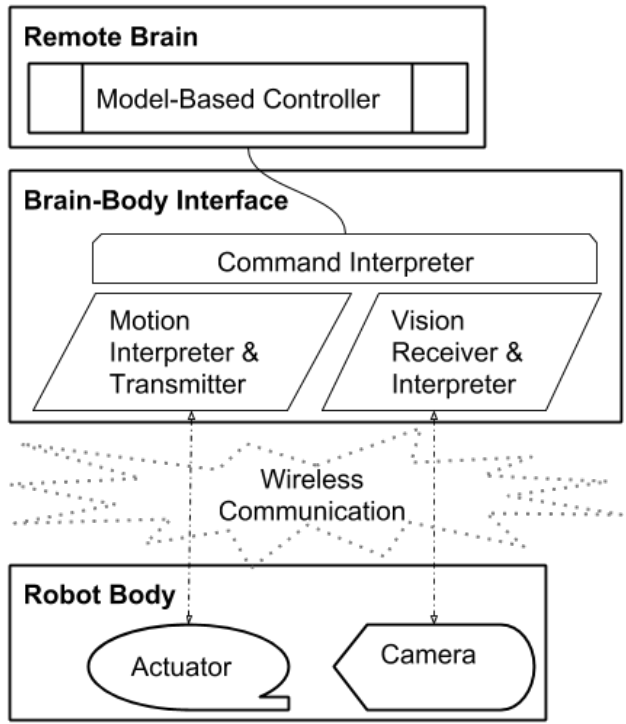

**Figure 1. The system architecture of an Apelike Robot system. (Adapted from [16])**

The Apelike robot can perform three main tasks: a) Model-based motion calculations, b) Vision-based interaction, and c) Vision-based stairs climbing. This multi-limbed robot is able to respond to dynamic visual feedback and perform some human-like actions. Although the robot was able to mimic the human behaviors, it still needs improvements to learn from the performed actions and develop more skills.

## 2.2 Remote-brained LEGO robot

In [25], authors presented an innovative NXT mobile robot system that can simulate the landmine detection operation with some assumption. The proposed system relies on moving an NXT mobile robot to develop a map for the environment. The robot mission is to scan an environment based on a developed grid to detect objects by using appropriate sensors attached to the mobile robot. The system was able to develop a map of the area under-study, showing the positions of located mines (i.e. objects). The early proposed remote-brained robot for object detection is presented in Figure 2.

The proposed system consists of three components. They are:

(i)     A PC Server is a Pentium 4 computer with a 512 RAM and support Wi-Fi connection. The PC Server works as the robot brain for the whole system since it is responsible of both monitoring and control of the system.

(ii)    The LEGO NXT is the mobile robot which is used to execute the mission commands. The NXT robot will hold the N80 mobile phone and navigate in the minefield.

(iii) N80 is a smart mobile phone device. N80 has an ARM-9 processor with clock rate of 220 MHz CPU, 18 MB RAM and Symbian OS v9.1[1]. N80 includes 3.0 Mega pixels camera and support for different types of wireless communication standards. They include (1) 3G network (384 kbps), (2) EDGE (286.8 kbps) (3) Bluetooth 2.0 and 4) WLAN 802.11b/g.

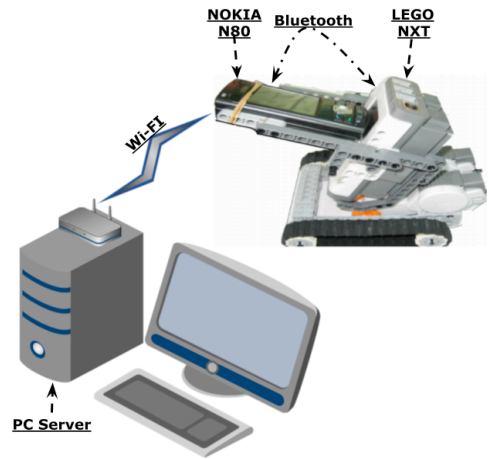

**Figure 2 Remote-brained LEGO Robot components (Adapted from [25])**

The operation of the proposed system is presented in Figure 3. In the proposed system, the N80 mobile works as a sensor camera that captures images, frequently. These images are sent to a PC server (i.e. the robot brain) such that a simple image processing algorithm is executed to detect objects in an image, if any. The result of execution is a command sent back via wireless communication to the robot to navigate in the specified environment.

The general model of the Remote-brained LEGO Robot is shown in Figure 4. It can be seen that the brain is in charge of analyzing the environment based on the collected data, the choice of the optimal decision to be taken based on available information and finally send appropriate command to robot actuators to complete a mission.

---

[1] Symbian OS is a proprietary operating system, designed for mobile devices, with associated libraries, user interface frameworks and reference implementations of common tools, produced by Symbian Ltd. It is a descendant of Psion's EPOC and runs exclusively on ARM processors.

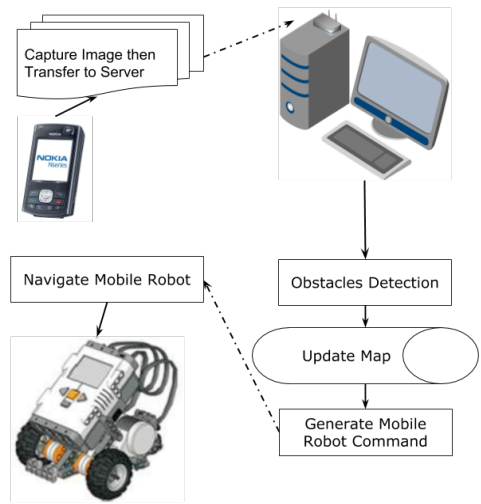

**Figure 3 Remote-brained LEGO Robot operations (Adapted from [25])**

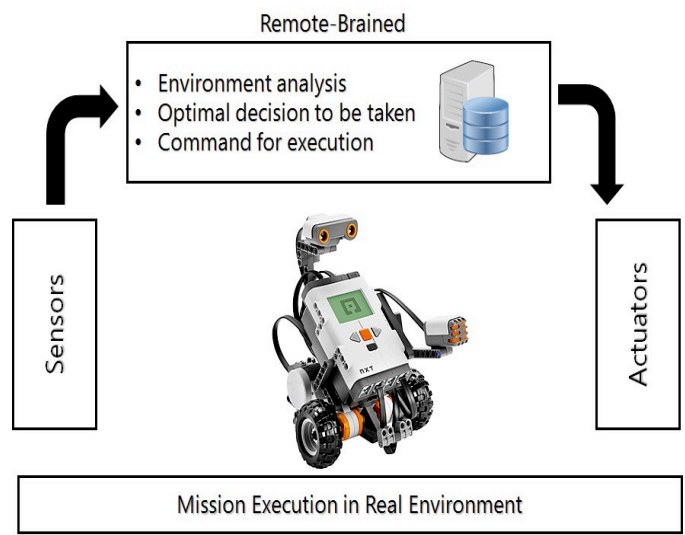

**Figure 4 The general model of the Remote-brained LEGO Robot**

## 3. Networked Robotics

In many applications such as search and rescue operation, a network of robots is always needed [28]. A networked robots system is defined as: "a group of robotic devices that are connected via a wired and/or wireless communication network" [29–31]. Networked robots allow multiple robots to overcome the restrictions of stand-alone robot by allowing robot sensors, brain and actuators to communicate via a wireless network or the World Wide Web (WWW) [32, 33]. This concept allows the development of mission with multi-robots with different types to collaborate in mission accomplishment such as searching for an object in an environment. Robots among the network can share and distribute tasks. Networked Robots allow the enhancement of interaction among robots, task integration, easy communication, and can enhance the overall robotics system security [31,34]. Part of research efforts in the area of networked robotics, related issues, and applications are presented in Table 2.

**Table 2 Part of Networked Robotics Research**

| Area/Subject | Objective | Ref. |
|---|---|---|
| Ubiquitous Networked Robotics | Studying re-usable and distributed applications as Ubiquitous Networked Robot Platform (UNR-PF). | [35] |
| Shared knowledge base | Integrating knowledge representation methods of the ROBOEARTH project with Ubiquitous Network Robot Platform. | [36] |
| Evaluation of challenges | Proposing and evaluating communication protocols, computing models, and discussing technical challenges. | [29] |
| Modular and cooperative robotics | Exploring tools and technologies that support modular and cooperative robotics. | [30] |
| … | | |

Authors in [35] presented the basic idea of a Ubiquitous Networked Robot Platform (UNR-PF). UNR-PF is introduced as a framework to coordinate and control distributed missions. The UNR-PF separated the hardware from the brain and uses a suitable interface that allows an application developers to create hardware- independent robotic services. In this case, a developer can demand components to fulfill a given need, and the UNR-PF will allocate suitable resources that can be controlled remotely. This approach helps in the design of re-usable distributed applications that can be used in various robotics platforms. It was found that networked robot architectures can facilitate the development, deployment, management and adaptation of distributed robotic applications [36]. A Networked Robots of NXT mobile robots with Internet and Machine-to-Machine communication are illustrated in Figure 5.

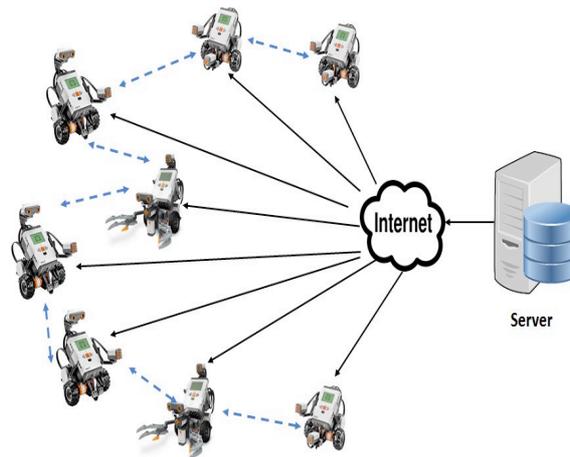

**Figure 5 Networked Robotics of NXT mobile robots with Internet and Machine-to- Machine (M2M) communication**

### 3.1 Problems with Networked Robotics

There were many reported challenges of the networked robots that stimulate the urge to develop an action plan for the coordination of the multiple autonomous robot systems in terms of communication, control, and perception. For example, which one should have the highest priority to establish communication? What is the maximum allowable time slot for each member of the network? How can each member of the networked robots access information? All these question could have a different answers according to time and place in the work environment. The development of a networked robotic system is always limited by resources, information, and communication constraints.

A summary of the main challenges faces the development of networked robotics is presented in [29] as follows:

(i) It is always the case that each robot member of the network has its own access to resources that include sensors, actuators and computing power (i.e. brain). The networked robotics resources can be accessed by any member of the network. The problem is once a robot is built and deployed, it is always not easy to change or upgrade their body structures such as the robot size, shape or power supply.

(ii) Networked robotics system is meant to develop a cooperative system that can accomplish a specific mission by using the available resources for each member of the network. Information fusion can be implemented in many ways such as machine-to-machine (M2M) or machine-to-server communication (M2S). Therefore, the networked robotics is restricted by the information sensed by the robots and shared over the network or the server program. This limits the ability of the network to learn.

(iii) Transferring information or decision making via a network involves the use of routing protocols. Routing protocols are responsible for sharing the infor- mation among the network members in a fast, efficient and reliable way. One of the known M2M communication protocols is the proactive routing. This protocol is usually used for routing messages across the network [37, 38]. Some of its disadvantages are: the need for high computation and memory resources to discover the best route over the network. The ad-hoc routing protocol is also used in networked robotics specially for dynamic routing [39, 40]. Ad-hoc routing protocols have number of problems related to establishing a path to send a message and overcome any related topology of a network. These sorts of challenges might tremendously affect the network performance and missions to be accomplished.

## 4. Cloud Computing

Cloud computing or on-demand computing, is a paradigm of Internet computing systems such that computing, memory and information are shared among set of devices belonging to a network. Cloud computing is considered as one of the most recent evolution in computing paradigms. The definition of cloud computing provided by the US National Institute of Standards and Technology (NIST) has gained significant attraction within the IT industry. According to this definition (see [NIST]):

> *Cloud computing is a model for enabling convenient, on-demand net- work access to a shared pool of configurable computing resources (e.g., networks, servers, storage, applications, and services) that can be rapidly provisioned and released with minimal management effort or service provider interaction.*

Cloud computing offers a wide range of elasticity and self-serving computing resources according to the system requirements [41]. It also aids in reducing infrastructure costs for many projects. Cloud computing establish a broad range of applications in variety of fields such as manufacturing, automation and control applications [42–45].

## 5. Cloud Robotics

The integration between cloud computing and robotics is found to be an upgrade to networked robotics. James J. Kuffner a Professor at Carnegie Mellon working at Google in 2010 was the first to coin the expression "Cloud Robotics" and described a number of potential benefits [46]. This concept allows the robot to become lighter, cheaper, and smarter. According to J. Kuffner, "robots that have access to a cloud could offload CPU-heavy jobs to remote servers, relying on smaller and less power-hungry on board computers. Even more promising, the robots could turn to cloud-based services to expand their capabilities." Later on, Steve Cousins introduced the concept of "No robot is an island".

The analysis of the scholarly articles, including patents, on GoogleScholar illustrates the emergence of four specific terms. Figure 6 illustrates the density of four specific search terms over several Years. Equation 1 is used to calculate the Term Density (TD) in a specific Year (y). Where Mentions is the total number of search results for a specific term in the current Year (y), and the (total) number of search results over all Years.

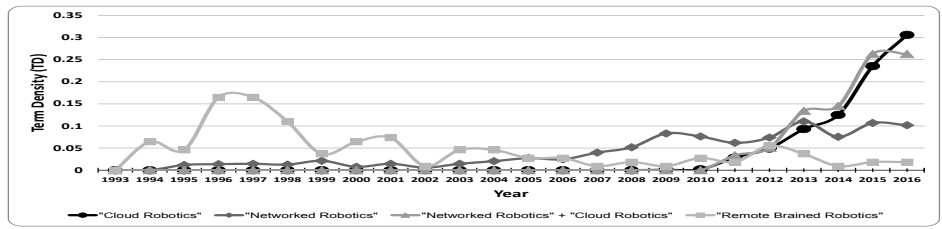

**Figure 6 Timeline of Terms Density based on GoogleScholar**

$$TD_y = \frac{Mentions_y}{Mentionsy_{total}} \times 100\% \qquad (1)$$

The search results count of the four terms "Cloud Robotic", "Networked Robotic", "Networked Robotics" + "Cloud Robotics", and "Remote Brained Robotics" till 23 August 2017 are 1160, 1180, 179, and 109 respectively. The numbers are based on the search query using GoogleScholar system which is prone to some errors. However, the results give clear indication about the emergence time of each search term and its level of attention among researchers over the Years. The existence of close relationship between cloud and networked robotics is supported by the analysis results.

Figure 7 illustrates an NXT Cloud Robot mobile system. The system consists of set of NXT robots that can connect to the cloud that act as a robot brain. Meanwhile, they can also communicate with each other. This allows the robots inside the system to get advantages of the accessibility to the cloud and the possible access to all robot's sensors inside the cloud.

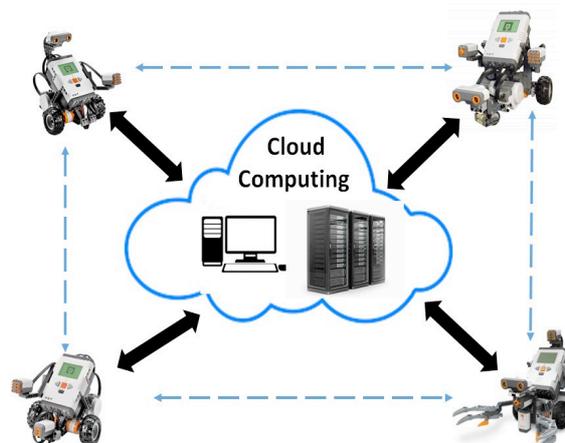

**Figure 7 NXT Cloud Robot mobile system**

Cloud computing was introduced as a solution to many constrained problems for networked robotics. These problems include limited computing power, unavailability of adequate storage space, and lack of communication and routing protocols. Networked robotics was considered as an evolutionary step on the way to cloud robotics [29]. A survey of research on cloud robotics and automation is presented in [47].

There are several up to date research efforts in the sake of formulating robust futuristic cloud-based robotic applications. Researchers are tackling various areas for an unlimited number of objectives to improve existing systems or to overcome critical limitations. Table 3 summarizes part of the most recent research efforts either in terms of real implementations or conceptual propositions. It is apparent that a large portion of the efforts presented in the table target industrial applications and large firms. Smart home, medical fields, education, and other domestic or small scale applications have been included as well.

**Table 3 Part of Most Recent Cloud-Robotics Research**

| Area/Subject | Objective | Ref. |
|---|---|---|
| Definition and insights | Definition of cloud robotics and its related issues. | [31] |
| Robotics & Automation as a Service (RAaaS) | Grasp-planning system, using Big Data and Data Mining | [48] |
| Industrial distributed apps | Performing surface blending using high-speed wide-area network | [49] |
| Conceptual Architecture | Manage Variability using domain-specific description language | [50] |
| Cooperative AGV systems in industrial warehouses | Cooperative data fusion system for global route assignment and local path planning | [51] |
| Ubiquitous manufacturing | Implementation and case studies to assess a developed framework and mechanisms | [52] |
| A review on frameworks | Review on Simultaneous Localization and Mapping | [53] |
| Cognitive Industrial IoT | Context-aware robotics for material handling | [54] |
| Service for robotic mental simulations | Mental simulation service for world simulation, learning algorithm, and solution suggestion using prolog queries | [55] |
| Smart Home | Enhancing off the shelf wireless robots | [56] |
| Switched reluctance machine in the direct-drive manipulator | Predictive current control to overcome latency and packet losses | [57] |
| Current state of cloud robotics | Review of up to date systems | [58] |
| Smart Manufacturing Environments | Review of main technologies in SMEs for self-adaptive adjustment, computing load allocation, and cloud-based group learning | [59] |
| Industrial Cloud Robotics | Case study to assess a proposed system related to energy conditions perception and Big Data analysis | [60] |
| Support of senior citizens | Testing technical capabilities of KuBo robot that is able to interact with humans and sense the environment. | [61] |

| Managing Internet of Drones | Implementation and validation of "Dronemap Planner" service. | [62] |
| Robot operating system | New protocol for robots operating system and Internet of Things (ROSLink) | [63] |
| Robot Cloud | Design and simulation of novel cloud service | [64] |
| Robots as a Service (RaaS) | Implementing and analyzing cloud robotics architecture. | [65] |
| Task Offloading | Implementing and evaluating Vehicular Cloud Computing system | [66] |
| Cyber-Physical Systems | Survey on remote brain, big data manipulation, and virtualization Robots | [67] |
| Risk Evaluation | Information security risk evaluation in Cyber Physical Systems | [68] |
| … | | |

### 5.1 Applications of Cloud Robotics

At the beginning, robots were designed and planned to be utilized as industrial robots with simple, repetitive and limited tasks [29, 69]. However, with the advancements of cloud robotic technologies, the robots are becoming lighter and smarter [41, 70, 71]. Due to this progress, in the last decade, there has been a real expansion in the applications of the cloud robotics [72, 73]. These applications cover a wide spectrum of applications and services in different domains including smart cities, transportation, health care and rehabilitation, housekeeping, environment monitoring, and entertainment [29, 47].

In cloud based robotics, robot units enjoy some powerful advantages like the accessibility to big data and shared knowledge, and the ability to transfer computation- intensive tasks to the high computing resources in the cloud [74]. These advantages enabled cloud robots to be deployed more efficiently in classical robotic applications such as simultaneous localization and mapping (SLAM) [75], grasping, and navigation [29]. In SLAM applications the robot localizes itself in unknown environment based on building maps for the environment. SLAM techniques are computationally and data intensive algorithms which makes could robots have a great advantage in such applications. An example of these techniques is Fast- SLAM which can be implemented in parallel processing in cloud frameworks [76]. Grasping is another common task in the field of robotics. This task can be computationally expensive and requires accessing huge amount of data when the targeted object is unknown. Moreover, deploying cloud robots for this task enables the robots to benefit from the shared training and learning experience. An example of this type of tasks can be found in [77] where a cloud robotics system was proposed for identifying and grasping common household objects based on Google Object Recognition Engine.

Different cloud robotics have been designed and deployed to support some services as a part of a specific application at different complexity levels. In this context, the term "robot-as-a-service" (RaaS) was coined to describe such robot units [78]. Examples of such cloud robot units are robot cops, robot waiters, robot pets, and patient and elderly care robots. Many of these applications depend heavily on the interaction and communicating with humans (known as human-robot inter-action (HRI))[69, 79]. Therefore, these units are based on observation (could be language, emotions recognition or even body expressions) and actions [69].

A worth mentioning example comes from Carnegie Mellon University and the Intel Pittsburgh Laboratory where a special service robot was designed, developed and named as Home Exploring Robotic Butler (HERB) where its main job is to care for the elderly and the disabled people [69, 80]. Such a service required an access to a vast amount of shared knowledge and experience collected from different robot units in order to cover and deal with a wide range of possible scenarios and interactions.

Another example can be found in [81] where the authors designed a cloud robot equipped with a camera capable of guiding impaired people to enjoy free tours in museums and live an experience by letting them see exactly what the robot is seeing in real-time. In general, cloud robot units could play different assigned roles as assistive technologies in schools, houses and offices [69, 82].

### 5.2 Algorithms for Cloud Robotics

Authors in [51] presented a data fusion system for optimizing the coordinated motion of industrial Automated Guided Vehicles (AGVs). They illustrated the contribution of two main algorithms in achieving better performance, namely "Geometric Level" and "Decision" data fusion algorithms. Geometric level algorithm enables autonomous global navigation and local path planning. On the other hand, decision algorithm is a "Global Live View" implementation at high level. Both algorithms enable processing sensors input to classify the obstacles in the environment. The authors show 93.2% overall classification rate in 15ms processing time of 60Hz "Global Live View". They claim that data fusion technologies and advanced sensing allow AGVs to overcome obstacles without the need for human intervention.

A genetic algorithm scheme tried to optimize "task offloading" for cloud robotics framework in [83]. The main objective of optimization was achieving optimal task decisions with minimal power consumption. A survey [47] covers an idea of the cloud as enabler for collective robot learning, crowd sourcing, open-source and open-access. The survey states that up to date, many robots still rely on their on-board computer, while there is a great opportunity in relocating the processing algorithms over computing clouds. In [64] the authors presented algorithms and an implementation of cloud-based system with the aim to leverage the cloud-based robot systems flexibility, reusability and extensibility.

In terms of architecture design, a domain specific language called (CRALA) enables designing and implementing cloud robotics architecture [84]. CRALA is considered an architecture description language that models three levels of architecture, namely specification, configuration, and assembly. Others [29, 85, 86] but not limited to proposed and illustrated various algorithms to support and improve the overall idea of cloud-based robot brain, including parallel processing approaches.

### 5.3 Challenges of Cloud Robotics

The authors in [87] discussed the key technical challenges of cloud robotics.

- First is the computation challenge that needs a common computation framework, optimizing the allocation and management of virtual resources, and bargaining between cloud and networked computation decisions.

- Second is the communication challenge which is prone to failure or delay in message passing, in addition to selecting the best topology of communication networks.

- Finally, the authors pointed to security and trust challenges regarding the cloud provider. Trust either when the robot tried to connect to the cloud or when executing a task over the cloud. Furthermore, storing data over the cloud might be with high risk, especially when the data is highly confidential or sensitive.

The quality of cloud services and network latency are expected to vary, therefore it is challenging to guarantee highly stable performance. The performance variability is critical for time sensitive applications that seek real-time responses [47, 88]. Due to tasks complexity in cloud-based robotic applications, it is challenging to select the best resource allocation and task scheduling approach. Apparently, communicating various structures of data from various sensors and devices creates a challenging interaction between the robots and the clouds. In terms of reliability analysis, there are no simple methods to assess the cloud-based robot system [88]. In [64] the authors suggested the need to tackle the challenges of reducing the running cost, improving the collaboration of robots to solve more complex tasks, and enhancing the adaptability of the robots in various domains.

## 6    Future Research

With the acceleration and advancement of cloud robotics and their applications, challenges become more sophisticated and complex. Challenges can fall into main streams like security, fault tolerance, scalability of the infrastructure, integration of affection and intelligence, quality of service provided by robot units, trust and the influence on interpersonal relationships.

Big data analytics tools for mining large data-sets captured by robots sensors is needed for improving human robot interactions, where this captured data (cloud data) will let researchers and practitioners in the field of cloud robots to design and develop new algorithms to enhance the knowledge of cloud robot systems [89]. Developing new algorithms for fault tolerance control, load balancing and for handling the varying in network latency are really open challenges in robots future research directions [90]. Moreover, scalable cloud infrastructures and data scaling techniques are needed where the first is important to scale the infra upon the size of the robots and the latter to scale the size of big data captured by robots to get better performance results.

On the other hand, ensuring effective privacy and security of data procedures is important since data that are captured by robots (cloud data) need to be secure by not allowing unauthorized party from accessing it or controlling the robot by the use of it [90]. In some critical cases, there is a need to contact a human to assist rather than a robot (available on-demand), where this approach must be balanced with the cost associated with it. Algorithms will be developed to decide when best to contact a human or when to proceed with the robot's control [47]. Optimistic researchers expect and demand developing robotic systems that fulfill real-time requirements [47, 88], and thorough analysis toward assuring the reliability and conformance to standards to achieve better performance [88].

In summary, cloud robotics is a result of radical advancements in the technology. In terms of computing, cloud services offer virtually unlimited scalability options. And in terms of communication technology, the wide spread of high speed wireless networks. However, as any generic cloud based service, security, privacy, communication cost, reliability, and many others are still with high level of concern. It is essential to assess the feasibility of cloud robotics in terms available technical capabilities, financial resources and running cost, and user acceptance. The idea of cloud robotics seems to be conceptually ideal in the meantime, but the concerns and open issues ask for investigating more in assurance mechanisms.

# Authors

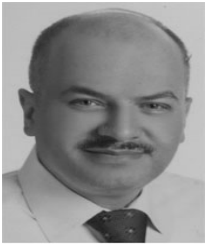

**Alaa F. Sheta,** is a Full time Assistant Professor/Department of Computing Sciences, Texas A&M University Corpus Christi, USA. He received his B.E., M.Sc. degrees in Electronics and Communication Engineering/Faculty of Engineering, Cairo University in 1988 and 1994, respectively. He received his Ph.D. from the Computer Science Department, George Mason University, USA in 1997. He published 150+ journal and conference papers, published two books, and co-editor of a book. His research interests include Evolutionary Computation, Software Reliability and Cost Estimation, Dynamical Nonlinear Systems Modeling & Simulation, Image Processing, Biotechnology, Business Intelligence, Robotics, Swarm Intelligence, Automatic Control, Fuzzy Logic & Neural Networks.

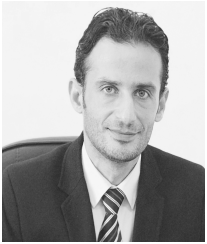

**Nazeeh Ghatasheh,** is an Associate professor at Business Information Technology Department/The University of Jordan (Jordan). Nazeeh Ghatasheh received his B.Sc. degree in Computer Information Systems from The University of Jordan, Amman, Jordan, in 2004. Then he was awarded merit-based scholarships to pursue his M.Sc. in e-Business Management and Ph.D. in e-Business at the University of Salento (Italy), that he obtained in 2008 and 2011 respectively. His research interests include e-Business, Business Analytics, Applied Computational Intelligence, and Data Mining. Ghatasheh, at present, is the Director of Computer Center at The University of Jordan, Aqaba, Jordan.

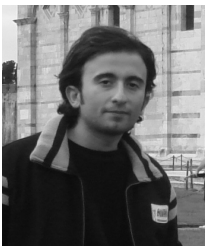

**Hossam Faris,** is an Associate professor at Business Information Technology Department/ The University of Jordan (Jordan). Hossam Faris received his B.Sc, M.Sc. degrees in Computer Science from Yarmouk University and Al-Balqa` Applied University in 2004 and 2008 respectively in Jordan. Then, he was awarded a scholarship to pursue his PhD degrees in e-Business at University of Salento, Italy, where he obtained his PhD degree in 2011. In 2016, he worked as a Postdoctoral researcher with GeNeura team at the University of Granada (Spain). His research interests include: Applied Computational Intelligence, Evolutionary Computation, and Data mining.

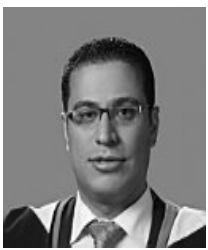

**Ali Rodan,** MIEEE (B.Sc. PSUT 2004; M.Sc. Oxford Brookes University 2005; Ph.D. The University of Birmingham 2012) is an Associate Professor of Computer science at the University of Jordan, Jordan. His research interests include recurrent neural networks, dynamical systems and machine learning. He has co-organized 4 special sessions devoted to learning on temporal data at international conferences. He is the Editor of a Book on Springer and a publisher of more than 10 JCR ISI journals. Currently he specializes on reservoir computation models including their design and theoretical properties.